\documentclass[runningheads]{llncs}

\usepackage{graphicx}
\usepackage{amsmath}
\usepackage{booktabs}
\usepackage{comment}
\usepackage{multirow}
\usepackage{makecell}
\usepackage{algorithm}
\usepackage[noend]{algpseudocode}
\usepackage{subcaption}
\usepackage{mathtools}
\usepackage{enumitem}
\usepackage{wrapfig}
\usepackage{url}
\usepackage[font=small,labelfont=bf,tableposition=top]{caption}
\usepackage[normalem]{ulem}  

\usepackage[capitalize]{cleveref}
\crefname{section}{Sec.}{Secs.}
\Crefname{section}{Section}{Sections}
\Crefname{table}{Table}{Tables}
\crefname{table}{Tab.}{Tabs.}

\usepackage{tikz}
\usepackage{comment}
\usepackage{amsmath,amssymb} 
\usepackage{color}

\usepackage[accsupp]{axessibility}  

\newcommand{\todo}[1]{}
\renewcommand{\todo}[1]{{\color{blue} {#1}}}

\usepackage{xspace}
\newcommand*{\eg}{e.g.,\@\xspace}
\newcommand*{\ie}{i.e.,\@\xspace}
\def\G{$\mathcal{G}$\@\xspace}
\def\Gaug{$\mathcal{G}_{aug}$\@\xspace}
\def\S{$\mathcal{S}$\@\xspace}

\DeclareMathOperator*{\argmin}{arg\,min}

\begin{document}
\pagestyle{headings}
\mainmatter
\def\ECCVSubNumber{6132}  

\title{Graph2Vid: Flow graph to Video Grounding for \\ Weakly-supervised Multi-Step Localization} 

\titlerunning{Flow graph to Video Grounding for Multi-Step Localization}
%
\author{Nikita Dvornik \and
Isma Hadji \and
Hai Pham \and
Dhaivat Bhatt \and
Brais Martinez \and
Afsaneh Fazly \and
Allan D. Jepson
}
\authorrunning{N. Dvornik, I. Hadji, H. Pham, D. Bhatt, B. Martinez, A. Fazly, A. Jepson}
%
\institute{Samsung AI Center}


\maketitle
\begin{abstract}
In this work, we consider the problem of weakly-supervised multi-step localization in instructional videos. 
An established approach to this problem is to rely on a given list of steps.
However, in reality, there is often more than one way to execute a procedure successfully, by following the set of steps in slightly varying orders.
Thus, for successful localization in a given video, recent works require the actual order of procedure steps in the video, to be provided by human annotators at both training and test times.
Instead, here, we only rely on generic procedural text that is not tied to a specific video.
We represent the various ways to complete the procedure by transforming the list of instructions into a procedure flow graph which captures the partial order of steps.
Using the flow graphs reduces both training and test time annotation requirements.
To this end, we introduce the new problem of flow graph to video grounding. 
In this setup, we seek the optimal step ordering consistent with the procedure flow graph and a given video.
To solve this problem, we propose a new algorithm - Graph2Vid - that infers the actual ordering of steps in the video and simultaneously localizes them. 
To show the advantage of our proposed formulation, we extend the CrossTask dataset with procedure flow graph information.
Our experiments show that Graph2Vid is both more efficient than the baselines and yields strong step localization results, without the need for step order annotation.

	
\end{abstract}

\section{Introduction}

Understanding video content from procedural activities has recently seen a surge in interest with various applications including future anticipation \cite{Sener2019,Girdhar2021}, procedure planning \cite{procedure2020,bi2021procedure}, question answering \cite{yang-hal-2021} and multi-step localization \cite{CrossTask,COIN,miech19howto100m,miech2020end,dropdtw}.
In this work, we tackle multi-step localization, \ie inferring the temporal location of procedure steps present in the video.
Since fully-supervised approaches \cite{caba2015activitynet,MaFK16,COIN} entail expensive labeling efforts, several recent works perform step localization with weak supervision.
The alignment-based approaches~\cite{richard2018neuralnetworkviterbi,ChangHS0N19,dropdtw} are of particular interest here as for each video they only require the knowledge of step order to yield framewise step localization.

However, all such alignment-based approaches share a common issue. They all assume that a given procedure follows a strict order, which is often not the case.
For example, in the task of making a pizza, one can either start with steps related to making dough, then steps involved in making the sauce, or vice-versa, before finally putting the two preparations together.
Since the general procedure (\eg recipe) does not define a unique order of steps, the alignment-based approaches rely on human annotations to provide the exact steps order for each video.
In other words, step localization via alignment requires using per-video step order annotations during inference, which limits the practical value of this setup.

\begin{figure}[t]
\centering
	\includegraphics[trim=0 150 140 0,clip,width=0.9\textwidth]{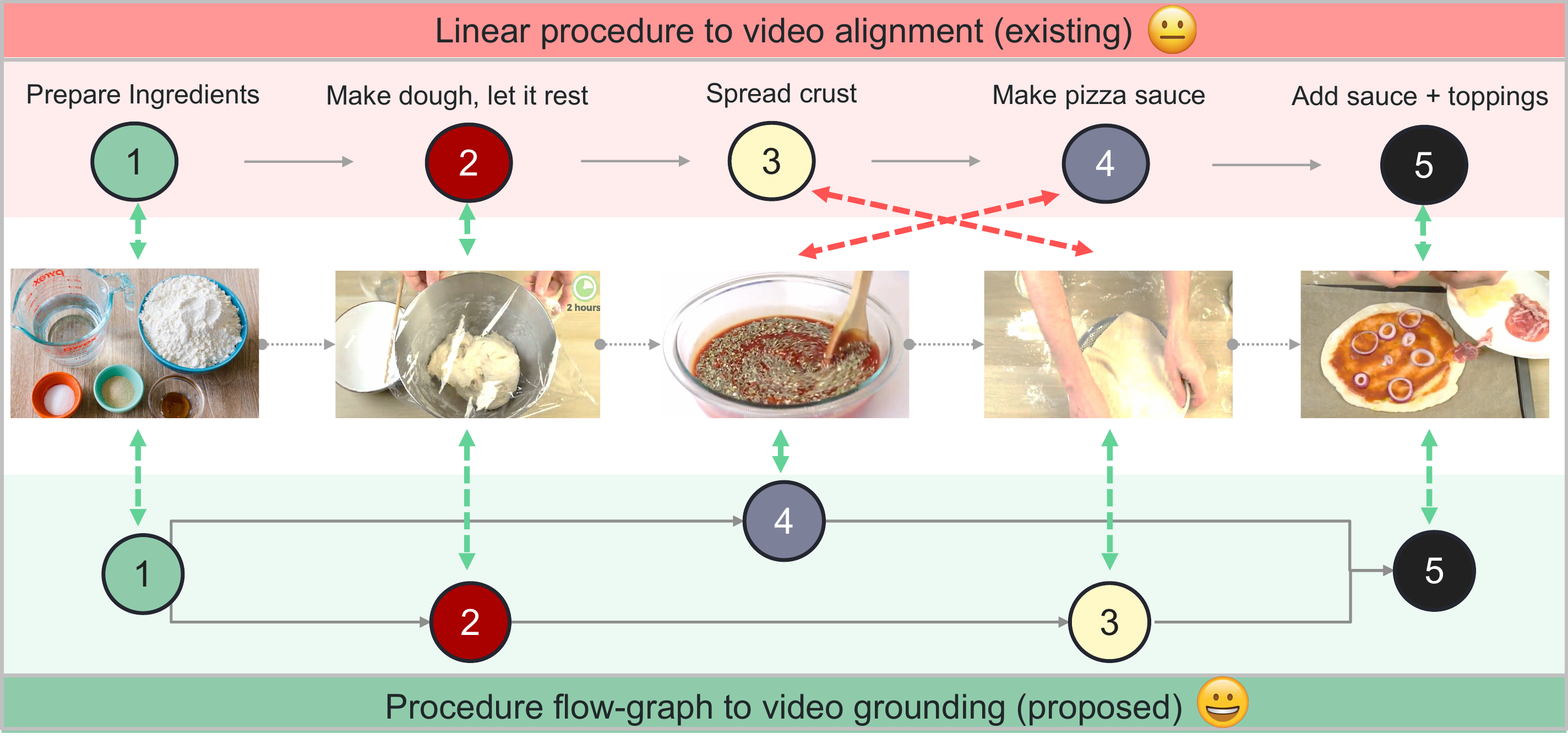}
	\caption{\textbf{Graph-to-Sequence Grounding.} (top) instructional videos do not always strictly follow a prototypical procedure order (\eg recipe). (bottom) Therefore, we propose a new setup where procedural text is parsed into a flow graph that is consequently grounded to the video to temporally localize all steps using our novel algorithm.} 
	\label{fig:teaser}
\end{figure}

To this end, we propose a new approach for step localization that does not rely on per-video step order annotation. 
Instead, it uses the general procedure description, common to all the videos of the same category (\eg the recipe of making pizza independent of the video sequence), to localize procedure steps present in any video.
Fig.~\ref{fig:teaser} illustrates the proposed problem setup. 
We propose to represent a procedure using a flow graph~\cite{schumacher2012-pk,Kiddon2015-ot}, \ie a graph-based procedure representation that encodes the partial order of instruction steps and captures all the feasible ways to execute a procedure.
This leads us to the novel problem of multi-step localization from instructional videos under the graph-based setting, which we call flow graph to video grounding.
To support the evaluation of our work we extend the widely used  CrossTask dataset \cite{CrossTask} with recipes and corresponding flow graphs.
Importantly, in this work, the flow graphs are obtained by parsing procedural text (\eg a recipe) freely available online using an off-the-shelf parser, which makes the annotation step automatic and reduces the amount of human annotation even further.


To achieve our goal of step localization from flow graphs, we introduce a novel solution for graph-to-sequence grounding - Graph2Vid. Graph2Vid is an algorithm that, given a video and a procedure flow graph, infers the temporal location of every instruction step such that the resulting step sequence is consistent with the procedure flow graph.
Our proposed solutions grounds each step in the video by: \textbf{(i)} expanding the original flow graph into a meta-graph, that concisely captures all topological sorts~\cite{topsort} of the original graph and
\textbf{(ii)} applying a novel graph-to-sequence alignment algorithm to find the best alignment path in the metagraph with the given video. 
Importantly, our alignment algorithm has the ability to ``drop" video frames from the alignment, in case there is not a good match among the graph nodes, which effectively models the no-action behavior.
Moreover, our Graph2Vid algorithm naturally admits a differentiable approximation and can be used as a loss function for training video representation using flow graph supervision.
As we show in Section 3, this can further improve step localization performance with flow graphs.

\noindent {\bf Contributions.} In summary the main contributions of this work are fourfold.
\begin{enumerate}
    \item We introduce flow graph to video grounding - a new task of multi-step localization in instructional videos given generic procedure flow graphs.
	\item We 
extend the CrossTask dataset by associating procedure text with each category and parsing the instructions into a procedure flow graph. 
	\item We propose a new graph-to-sequence grounding algorithm (\ie Graph2Vid) and 
	show that Graph2Vid outperforms baseline approaches in step localization performance and  efficiency. 
	\item We show Graph2Vid can be used as a loss function to supervise video representations with flow graphs.
	
\end{enumerate}

The code will be made available at \url{github.com/SamsungLabs/Graph2Vid}.
\section{Related work}
\noindent{\bf Sequence-to-sequence alignment.}
Sequence alignment has recently seen growing interest across various tasks  \cite{tcn,softdtw,ChangHS0N19,DwibediATSZ19,d2tw,CaoJCCN20,Chang2021,CaiXYHR19}, in particular, the methods seeking global alignment between sequences by
relying on Dynamic Time Warping (DTW) \cite{DTW,softdtw,ChangHS0N19,d2tw,Chang2021}. Some of these methods 
propose differentiable approximations of the discrete operations (\ie the min operator) to enable training with DTW~\cite{softdtw,d2tw}. 
Others, allow DTW to handle outliers in the sequences~\cite{MullerMeindard,CaoJCCN20,SakuraiFY07,NW,SmithWaterman,LCSS}. Of particular note, the recently proposed Drop-DTW algorithm~\cite{dropdtw} combines the benefits of all those methods as it allows dropping outliers occurring anywhere in the sequence, while still maintaining the ability of DTW to do one-to-many matching and enabling differentiable variations.
However, as most other sequence alignment algorithms, Drop-DTW matches sequence elements with each other in a linear order, not consider possible element permutations within each sequence.
In this work, we propose to extend Drop-DTW to work with partially ordered sequences. This is achieved by representing one of the sequences as a directed cyclic graph, thereby relaxing the strict order requirement.

\noindent{\bf Graph-to-sequence alignment.}
Aligning graphs to sequences is an important topic in computer science. One of the pioneering works in this area proposed a Dynamic Programming (DP) based solution for pattern matching where the target text is represented as a graph~\cite{NAVARRO2000}. Many follow up works extend this original idea via enhancing the alignment procedure. Examples include, admitting additional dimensions in the DP tables for each alternative path \cite{POA}, improving the efficiency of the alignment algorithm \cite{Mikko2019,Jain2019OnTC} or explicitly allowing gaps in the alignment, thereby achieving sub-sequence to graph matching \cite{Kavya2019SequenceAO}.
A common limitation among all these methods is the assumption that only \emph{one} of the paths in the graph aligns to the query sequence, while alternative paths and their corresponding nodes do not appear in the query sequence. Therefore, the goal in graph-to-sequence \emph{alignment} is to find the specific path that best aligns with the query sequence. 
In contrast, we consider the novel problem of graph-to-sequence \emph{grounding}. In particular, we consider the task where all nodes in the graph have a match in the query sequence and therefore our task is to ground each node in the sequence, while finding the optimal traversal in the graph that best aligns with the sequence.  
This problem is strictly harder than graph-to-sequence alignment and can not be readily tackled by the existing algorithms.

\noindent{\bf Video multi-step localization}
The task of video multi-step localization has gained a lot of attention in the recent years \cite{miech2020end,Luo2020UniVL,dropdtw} particularly thanks to instructional videos dataset availability that support this research area \cite{CrossTask,COIN,YouCook2}. The task consists of determining the start and end times
of all steps present in the video, based on a description of the procedure depicted in the video. Some methods rely on full supervision using fine-grained labels indicating start and end times of each step (\eg \cite{caba2015activitynet,MaFK16,COIN}). However, these methods require extensive labeling efforts. Instead, other methods propose weakly supervised approaches where only steps order information is needed to yield framewise step localization~\cite{HuangFN16,DingX18,richard2018neuralnetworkviterbi,ChangHS0N19,CrossTask,dropdtw}. However, these methods lack flexibility as they require \emph{exact} order information to solve the step localization task.
Here, we propose a more flexible approach where only partial order information, as given by a procedure flow graph, is required to localize each step.
In particular, given a procedure flow graph, describing all possible step permutations that result in successful procedure execution, our method localizes the steps in a given video, by automatically grounding the steps of the graph in the video.

\section{Our approach}
\vspace{-5pt}
In this section, we describe our approach for flow graph 
to video grounding. We start with a motivation and formal definition of our proposed flow graph to sequence grounding problem. Next, we describe in detail our proposed solution to tackle the task of video multi-step localization using flow graphs.
\subsection{Background}\label{sec:background}
\textbf{Ordered steps to video alignment.}
If the true order of steps in a video (\ie as they happen in a video) is given, the task of step grounding reduces to a well-defined problem of steps-to-video alignment, which can be solved with some existing sequence alignment method.
In particular, the recent Drop-DTW~\cite{dropdtw} algorithm suits the task particularly well thanks to a unique set of desired properties: \textbf{(i)} it operates on sequences of continuous vectors (such as video and step embeddings) \textbf{(ii)} it permits one-to-many matching, allowing multiple video frames to match to a single step
, and \textbf{(iii)} it allows for 
dropping elements from the sequence, 
which in turn allows for ignoring video frames that are unrelated to the sequence of steps. In Drop-DTW, the alignment is formulated as minimization of the total match cost between the video clips and instruction steps. It is solved using dynamic programming and can be made differentiable (see Alg.~1 in~\cite{dropdtw}).
That is, given a video, $\mathbf{x}$, and a sequence of steps, $\mathbf{v}$, Drop-DTW returns the alignment cost, $c^\ast$, and alignment matrix, $M^\ast$, indicating the correspondences between steps and video segments.\\ 

\noindent \textbf{Procedure flow graphs.} In more realistic settings, procedure steps for many processes, such as cooking recipes, are often given as a set of steps in a partial order. Specifically, the partial ordering dictates that certain steps need to be completed before other steps are started, but that other subsets of steps can be done in any order.
For example, when thinking of making a salad, one can cut tomatoes and cucumbers in one order or the other, however we are certain that both ingredients must be cut before mixing them into the salad. 
This is an example of a procedure with partially ordered steps; \ie there are multiple valid ways to complete the procedure, all of which can be conveniently represented with a flow graph.

\begin{figure}[t]
\centering
\includegraphics[trim=0 190 11 0,clip,width=0.99\textwidth]{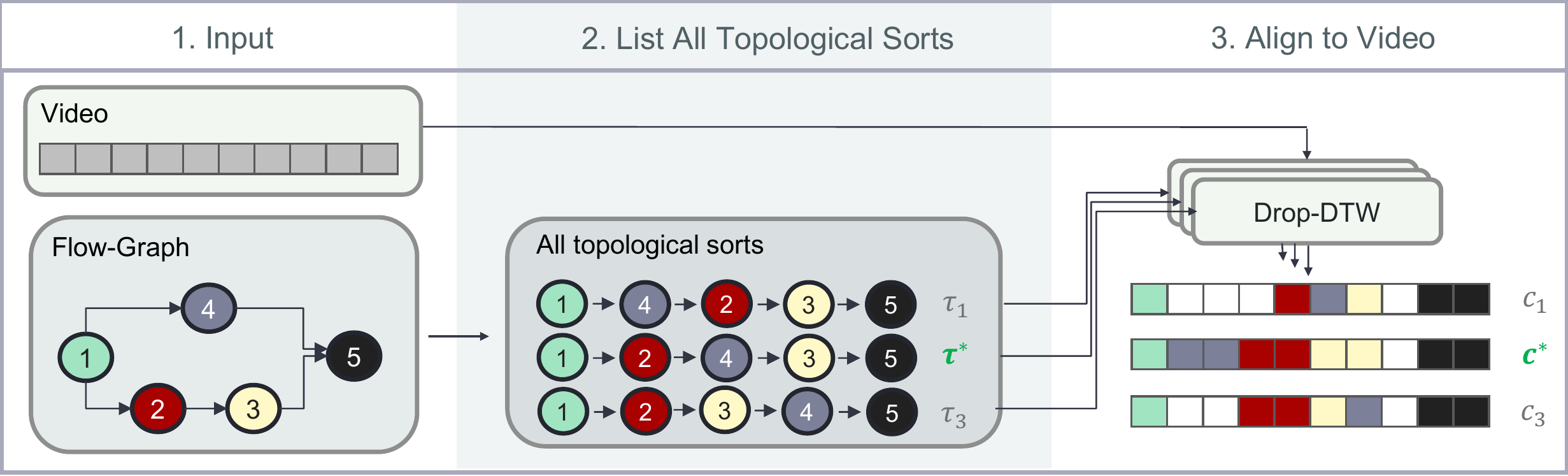}
\caption{\textbf{Brute-force approach for flow-graph to video grounding.}
Given the the flow-graph, the brute-force approach explicitly considers every topological sort and aligns it to the video with Drop-DTW independently, making the inference inefficient.
\label{fig:bruteforce}
}
\end{figure}

A procedure flow graph is a Directed Acyclic Graph (DAG) $\mathcal{G}=(V, E)$, where $V$ is a set of nodes and $E$ is a set of directed edges. Each node $v_i \in V$ represents a procedure step and an edge $e_j \in E$ connecting $v_k$ and $v_l$ declares that the procedure step $v_k$ must be completed before $v_l$ begins in any instruction execution. If a node $v_k$ has multiple ancestors, all the corresponding steps must be completed before beginning instruction step $v_k$.
In this work, we assume that \G has a single root and sink nodes. For convenience, we automatically add them to the graph if they are not already present.
From the definition of the flow graph, it follows that every topological sort~\cite{topsort} of the nodes in \G (see Fig.~\ref{fig:bruteforce}, step 2) is a valid way to complete the procedure.
This is an important property that forms the foundation of our Graph2Vid approach, described next.\\

\noindent \textbf{Flow graphs to video grounding.}
We define the task of grounding a flow graph $\mathcal{G}$ in a video $\mathbf{x} = [x_i]_{i=1}^N$, where $N$ is the total number of frames, as the task of finding a disjoint set of corresponding video segments, $s_l = [x_i]_{i=start_l}^{end_l}$ for each node $v_l \in \mbox{\G}$ of the flow graph, such that the resulting segmentation conforms to the flow graph. Specifically, in a pair of resulting video segments, ($s_i$, $s_j$), segment $s_i$ can only occur before $s_j$ in the video if the corresponding node $n_i$ is a predecessor of $n_j$ in the flow graph $\mathcal{G}$.
In this work, we assume that every procedure step $v_l$ appears in the video exactly once.

\subsection{Graph to video grounding - a brute-force approach}\label{sec:naive}
Before introducing our method, we first discuss a naive solution to the problem. As previously mentioned, we know that the procedure execution in the video follows some topological sort of the flow graph, \G. Thus, one can derive an algorithm for flow graph to video grounding by explicitly considering all the topological sorts of \G as depicted in Fig.~\ref{fig:bruteforce}.
More specifically, one can generate all topological sorts of the flow graph, try aligning each of them to the video, and select the best alignment as step grounding.
This essentially reduces the problem of flow graph grounding to sequence-to-sequence alignment which can be solved using Drop-DTW. The procedure would iterate over all topological sorts $\tau \in \mathcal{T}$ of the flow graph $\mathcal{G}$ and align a sequence of nodes (re-ordered according to the topological sort) $\mathbf{v^{\tau}} = [v_{\tau_{i}}]_{i=1}^{|V|}$ to the video sequence, and pick the alignment $M^\ast$ with the topological sort $\tau^\ast$ that has the minimum alignment cost $c^\ast$:
\begin{equation}
    c^\ast = \texttt{Drop-DTW}(\mathbf{v^{\tau^\ast}}, \mathbf{x}); \quad \tau^\ast = \argmin_{\tau \in \mathcal{T}} \texttt{Drop-DTW}(\mathbf{v^{\tau}}, \mathbf{x})
\end{equation}
While simple, this approach has one crucial downside - its efficiency.
As we show in Sec.~\ref{sec:complexity}, this algorithm has exponential complexity and becomes infeasible even for small flow graphs.
Thus, we need a more scalable solution to the problem.

\subsection{Graph2Vid - our approach}
\begin{figure}[t]
\centering
\includegraphics[trim=0 185 25 0,clip,width=0.99\textwidth]{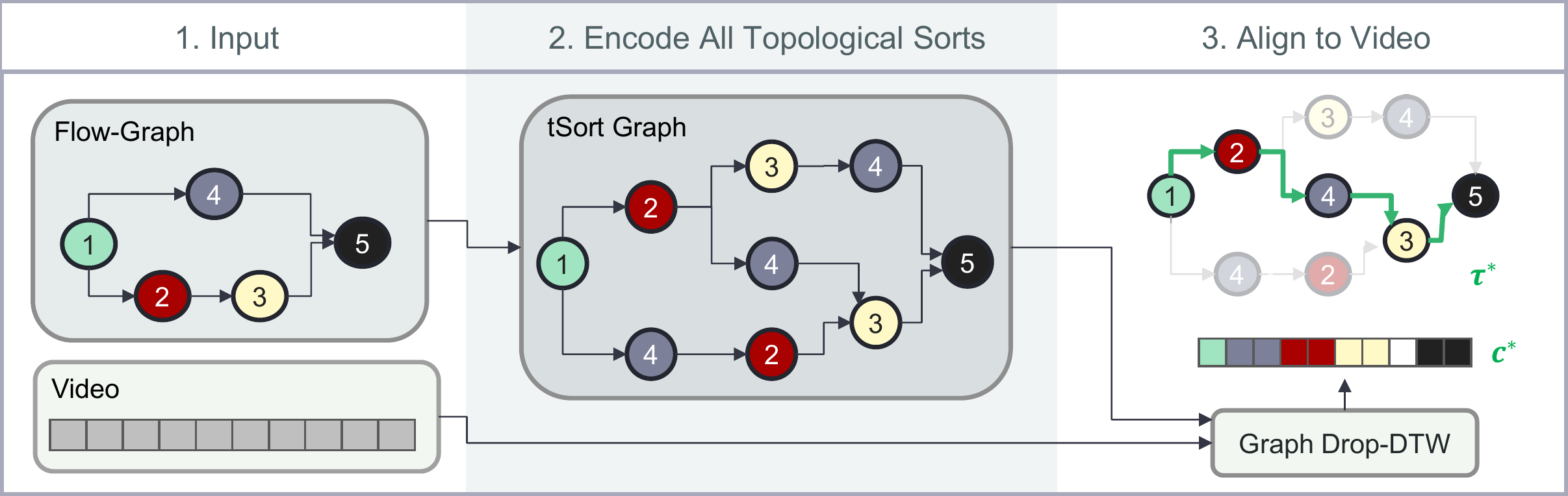}
\caption{\textbf{Flow-graph to video grounding with Graph2Vid.}
Given the the procedure flow-graph and a video as input our method packs all topological sorts into a tSort Graph and then uses Graph-Drop-DTW to align it to the given video, producing video segmentation and the optimal topological sort $\tau^\ast$.
\label{fig:graph2vid}
}
\end{figure}

In this section we present a new, more efficient approach for flow graph to video grounding, that we term Graph2Vid (see the overview in Fig.~\ref{fig:graph2vid}).
Graph2Vid operates in two stages: \textbf{(i)} given the flow graph, \G, we pack all the topological sorts of \G into a novel compact graph-based representation \S (that we call the tSort graph); \textbf{(ii)} we align the obtained tSort graph, \S, to the video, $\mathbf{x}$, using a proposed graph-to-video alignment algorithm, which we dub Graph-Drop-DTW. To compute the alignment we embed the procedure text in each node of the graph, $v_\tau$, and the clips of the video sequence, $x_i$, using a model pre-trained on the large HowTo100M dataset \cite{miech2020end}. The key to superior efficiency of Graph2Vid is the interplay between the tSort graph structure and the graph-to-video alignment algorithm, which allows for polynomial complexity of flow graph to video grounding\footnote{The tSort graph is polynomial for an assumed subset of flow graphs with a fixed maximum number of threads.}.
Moreover, Graph2Vid allows for a differentiable approximation that can be used for training neural networks.
In the following, we explain how to construct the tSort graph, \S, from the original flow graph, \G, how to extend Drop-DTW to perform graph-to-sequence alignment, and finally, how to use Graph2Vid as a differentiable loss function.\\

\noindent \textbf{Creating the tSort graph.} As we have shown previously, grounding a flow-graph, \G, to a video requires considering all the topological sorts of \G, yet their explicit consideration is infeasible due to the exponential number of topological sorts.
We note that the cause of such inefficiency is the redundancy and high overlaps between the topological sorts.
This motivates us to encode all topological sorts into a tSort graph, \S, that effectively shares the common parts of different topological sorts and provides a more compact representation.
Fig.~\ref{fig:graph2vid} illustration how for a simple flow graph input we construct the tSort graph encoding all the topological sorts (Fig.~\ref{fig:bruteforce}).
Each path from root to sink in the tSort graph, \S, spells out a topological sort of \G. Thus, listing all
root to sink paths in
\S is equivalent to listing all topological sorts of the original flow graph, \G.\\
Algorithm~\ref{algo:tsort} gives a procedure for constructing the tSort graph \S.
The first step to constructing the tSort graph is to connect all the nodes on separate threads in the original flow-graph \G (\ie like nodes $\{4\}$ and $\{2, 3\}$ in Fig~\ref{fig:graph2vid}) with undirected edges.
Since the instruction steps on separate threads may follow one after another in an actual instruction execution, thus connection between them must exist. These connections yield an augmented graph, \Gaug.
Then, in this augmented graph \Gaug, we run Breat First Search (BFS) traversal to find all the paths that lead from the root to the sink node of \Gaug and conform to the original flow-graph \G.
During the BFS traversals, the paths that have visited the same set of nodes so far are merged together and mapped into a single node in the tSort graph \S.
Merging the traversals with the same set of visited nodes enables the tSort graph to represent all topological sorts efficiently.
For more details on the tSort graph construction and a more efficient implementation description, we refer the reader to Appendix.

\begin{algorithm}
\small
	\caption{tSort-graph Construction\label{algo:tsort}} 
	\begin{algorithmic}[1]
		\State \textbf{Inputs}: \G - flow graph, $s$ - root node in \G
         \State \Gaug = aug(\G) \label{alg:tsort_aug}
		 \Comment{\textit{connect the nodes on parallel threads}}
		 
         \State $E_{tSort}$ = []
		 \Comment{\textit{init edge set of the tSort-graph}}
		 
		 \State $q$ = queue((s, $\oslash$))
		 \Comment{\textit{init BFS queue}}
		 
		 \While{q}
		 \State $v, P$ = $q$.pop() \Comment{\textit{active node v, set of visited nodes P}}
		 \For{$v_d$ in get\_descendants($v$, \Gaug)}
		 \State $P_d$ = $P$.add($v_d$)
		 \Comment{\textit{extend the visited nodes set}}
		 
		 \If{get\_predecessors($v_d$, \G) in $P$} \label{alg:tsort_conform}
		 \Comment{\textit{the path $P_d$ conforms to \G}}
		 \State $q$.append(($v_d$, $P_d$))
		 \State $E_{tSort}$.add((($v, P$), ($v_d, P_d$))) \label{alg:tsort_addedge}
		 \Comment{\textit{add edge to tSort graph}}
		 
		 \EndIf \EndFor \EndWhile
        \State \S = build\_graph\_from\_edges($E_{tSort}$)
		 \Comment{\textit{build the tSort graph}}
        \State  \textbf{Output:} \S
	\end{algorithmic}
\end{algorithm}

\noindent\textbf{Graph to sequence alignment using Graph-Drop-DTW.}
Having access to the tSort graph, \S, (whose every path from root to sink is a valid topological sort of the flow graph, \G), we can cast flow graph to video grounding as graph-to-video alignment problem~\cite{NAVARRO2000}.
In this case, the graph-to-video alignment finds a traversal of the graph \S that best aligns with the given video $\mathbf{x}$.
Importantly, directly aligning the tSort graph to a video discovers the optimal path in the graph and the best alignment of this path to the video \emph{simultaneously} (see Fig~\ref{fig:graph2vid}, step 3). This is in contrast to the naive solution in Sec.~\ref{sec:naive}, which uses sequence alignment as a subroutine to find the optimal topological sort.

In order to solve graph to video grounding, the graph-to-sequence alignment algorithm must have the following properties:  \textbf{(i)} operate on continuous vectors, \textbf{(ii)} permit one-to-many matching and \textbf{(iii)} allow for unmatched sequence elements.
To the best of our knowledge, a graph-to-sequence alignment algorithm with such properties does not exist.
However, Drop-DTW~\cite{dropdtw} for sequence alignment satisfies all 3 criteria (see Sec.~\ref{sec:background})
Thus, we propose a new algorithm - Graph-Drop-DTW - which is an extension of Drop-DTW for graph-to-sequence alignment. 
We base Graph-Drop-DTW on Alg.1 in~\cite{dropdtw} and modify the dynamic programming recursion to take into account the graph structure as follows:

\begin{eqnarray}\label{eq:graph_drop_dtw}
    D^+_{i,j} &=&  C_{i,j} +  \min \{\min_{k \in A(i)}\{D_{k,j-1}\}, D_{i,j-1}\}  \\
    D^-_{i,j} &=& d^x_{j} + D_{i,j-1} \nonumber \\
    D_{i, j} &=& \min\{D^+_{i, j}, D^-_{i, j}\}, \nonumber
\end{eqnarray}
where $C_{i, j}$ is the cost of matching the $i$-th node of the graph to the $j$-th video clip, and $d^x_j$ is the cost of dropping the $j$-th video clip and not matching it to any node in the graph.
We follow~\cite{dropdtw} and define $C_{i,j}$ as negative log-likelihood of the video clip $j$ belonging to step $i$, and compute the drop cost $d^x_j$ as a 30-th bottom percentile of $C_{i,j}$.
Different from Drop-DTW, when computing $D^+_{i,j}$, Graph-Drop-DTW takes into consideration all the predecessors of the node $i$, (\ie $k \in A(i)$) and selects the one minimizing the alignment cost.
Intuitively, the minimum operation over the predecessors, $\min_{k \in A(i)}$, in Eq.~\eqref{eq:graph_drop_dtw}, dynamically finds the traversal of the graph that aligns with the video best.
Similar to Drop-DTW, Graph-Drop-DTW outputs the alignment cost $c^\ast$ and the alignment path $M^\ast$, representing the optimal matching between the nodes of the input graph \S and the video clips of $x_i \in \mathbf{x}$.
It is important to note that Graph-Drop-DTW can only drop elements from the sequence (as a direct extension of Alg.1 in~\cite{dropdtw}) and does not support dropping nodes from the graph.
\\

\noindent\textbf{Graph2Vid for flow graph grounding}
Finally, Graph2Vid can be defined as the complete pipeline that chains tSort graph creation and its alignment to the video. Precisely, given the procedure flow graph, \G, and the video sequence, $\mathbf{x}$, Graph2Vid first transforms \G into the tSort graph, \S, then aligns the graph, \S, to the video, $\mathbf{x}$, using Graph-Drop-DTW.
This effectively provides the desired correspondence between every node, $v_i \in \mbox{\G}$, and a video segment in $\mathbf{x}$, that conforms to the flow graph, \G. 

\subsection{Graph2Vid for representation learning}
We now describe a differentiable approximation of Graph2Vid to learn video representations using flow graphs as the source of supervision.
To use Graph2Vid as a loss function, we must be able to backpropagate gradients with respect to the video input. That is, Graph-Drop-DTW must be differentiable.
A differentiable version of Graph-Drop-DTW can be obtained by simply using a soft approximation of the $\min$ operator (\eg \cite{softdtw,d2tw}) in Eq.~\eqref{eq:graph_drop_dtw}.
Here, we substitute the $\min$ in Eq~\ref{eq:graph_drop_dtw} with the smooth-min operator~\cite{d2tw} defined as
\begin{equation}\label{eq:soft_dropdtw}
    hey
\end{equation}
where $\gamma > 0$ is a hyper-parameter controlling the trade-off between smoothness and the error of the approximation.

With this differentiable version of Graph2Vid, we can use the matching cost between a video and it's corresponding flow graph as a training signal. Intuitively, the lower the cost of matching of a video to its corresponding procedure flow graph, the better are the learned representations.
Specifically, we define the Graph-Drop-DTW loss as the cost of grounding the flow graph, \G, to video, $\mathbf{x}$:
\begin{equation}
    \mathcal{L}_{\text{G}}(Z, \mathbf{x}) = c^\ast = \textrm{Graph-Drop-DTW}( \mathcal{G}, \mathbf{x}) \label{eq:loss-Graph-Drop-DTW}.
\end{equation}
As we show in Sec.~\ref{sec:experiments}, using $\mathcal{L}_{\text{G}}$ for weakly-supervised learning (with flow-graph supervision) improves step localization performance.

\subsection{On the algorithm's complexity}\label{sec:complexity}
To develop some intuition on both the size of the generated tSort-graphs, and on the speed-up over the naive approach, we consider simple model problems where the flow graph, \G, consists of $T$ separate, linearly-ordered threads, with $n_1, n_2, \ldots, n_T \geq 1$ nodes in each thread, for a total of $\sum_{t=1}^T n_t = n$ steps.  
As we show in Appendix, in this case, the number of topological sorts of \G, $N_{ts}(\mathcal{G})$ grows exponentially with $n$ and has the complexity $O(T^n)$.
On the other hand, the number of nodes in the tSort graph \S, $|V_S|$, is polynomial in the number of nodes $n$, \ie $|V_S| = O(Tn^T)$ which is better than exponential growth in $N_{ts}(\mathcal{G})$, provided that $T$ is not too large.
As shown in the Appendix, the asymptotic speedup of Graph2Vid over the brute-force approach can be roughly described by the ratio $N_{ts}(\mathcal{G}) / |V_s|$ which is still exponential in $n$, giving a large advantage to Graph2Vid.
This is further confirmed in Sec.~\ref{sec:ablation}, where Graph2Vid is orders of magnitude faster than the brute-force approach on real procedure flow-graphs.

\subsection{Creating flow graph from procedural text}\label{sec:parser}

To construct flow graphs from procedural text,we considered two automated alternatives; namely, a rule-based and a learning-based approach. In addition, we considered a manual approach, where we explicitly define nodes and edges.\\

\noindent \textbf{Rule-based graph parsing.}
Starting from regular procedural instructions, we first extract relevant text entities from each step description, including action verbs, direct and prepositional objects as suggested in~\cite{schumacher2012-pk}. Next, we define a set of rich semantic rules for the graph constructor to connect the text entities. Once edges are defined using those entities, flow graphs are collapsed into coarse sentence-level graphs to be used in our experiments.\\

\noindent\textbf{Learning-based graph parsing.} Using the same procedural instructions, we consider a learning based approach, which relies on two steps. First, we identify $10$ named entities using the the tagger of~\cite{yamakata2020}. Second, we used the parser proposed recently in~\cite{donatelli2021} to automatically find edges between the named entities defined in the first step. Once again, we finally collapse the fine-grained flow graphs into coarse sentence-level graphs as done with the rule-based parser.
\\
A detailed description of both approaches is in Appendix.

\section{Experiments}\label{sec:experiments}
To demonstrate the strengths of the proposed Graph2Vid algorithm, we first present our dataset construction in Sec.~\ref{sec:Datasets}. Then we describe the metrics used to evaluate the task of multi-step localization from graphs as well as the adopted baselines in Sec.~\ref{sec:metrics-baselines}. Finally, we summarize our results in Secs.~\ref{sec:results} through~\ref{sec:ablation}.

\subsection{Dataset construction}\label{sec:Datasets}
To evaluate our new formulation of graph-to-video grounding using the proposed Graph2Vid approach, we need a dataset with procedure steps captured in flow graphs and corresponding Ground Truth (GT) start and end times for each node in these graphs. To this end, we extend the widely used CrossTask datast \cite{CrossTask}  following three main steps: \textbf{(i)} For each procedure class (\eg \textit{Build floating shelves} or \textit{Making pancakes}), we grab the procedure text from the web~\footnote{We find the procedure text of CrossTask in www.wikihow.com}. \textbf{(ii)} We extract flow graphs from the procedural text following the methods described in Section~\ref{sec:parser}, such that each node in the flow graph corresponds to a step from the procedure text. \textbf{(iii)} Finally, we manually find correspondence between nodes in the graph and step instructions provided with the original datasets. These correspondences are used to associate the original GT temporal annotations of each step with nodes in our flow graph. These GT temporal annotations, now associated with our graph nodes, are used to evaluate our model on the task of multi-step localization. Importantly, these annotations are only necessary for evaluation, and not for training or inference.

\subsection{Metrics and baselines}\label{sec:metrics-baselines}
We evaluate the performance of our Graph2Vid approach on multi-step localization using two different metrics: \textbf{(i) Framewise accuracy (Acc.)}~\cite{COIN}, which is defined as the ratio between the number of frames assigned the correct step label (not including background) and the total number of frames, and \textbf{(ii) Intersection over Union (IoU)}~\cite{YouCook2}, which is defined as the sum of the intersections between the predicted and ground truth time intervals for each step label divided by the sum of their unions.

As we are the first to tackle multi-step localization under this new paradigm of flow graph to video grounding, we consider three increasingly strong baselines for comparisons. \textbf{(i) Bag of steps.} In this baseline, we consider every steps as being a node in a separate thread in a graph (\ie no graph structure or order is imposed). \textbf{(ii) Linear Procedure.} Here, we read the instructions extracted from the procedure text linearly and assume this order as the default ordering of steps. \textbf{(iii)} \textbf{GT Step Sequence.} This is the \emph{upper bound} on our Graph2Vid; it uses the ground truth step order (\ie as they happen in each video) provided with the dataset. For baseline (i) we use the proposed Graph-Drop-DTW to obtain a segmentation as we treat the bag of steps as a disconnected graph. On the other hand, in baselines (ii) and (iii), we use the Drop-DTW algorithm \cite{dropdtw} to obtain the segmentations, which we chose for two main reasons. First, this algorithm directly relates to the proposed Graph-Drop-DTW. Second, Drop-DTW is currently state-of-the-art on CrossTask for step localization~\cite{dropdtw}. 


\begin{figure}[t]
\centering
	\includegraphics[trim=0 190 15 0,clip,width=0.99\textwidth]{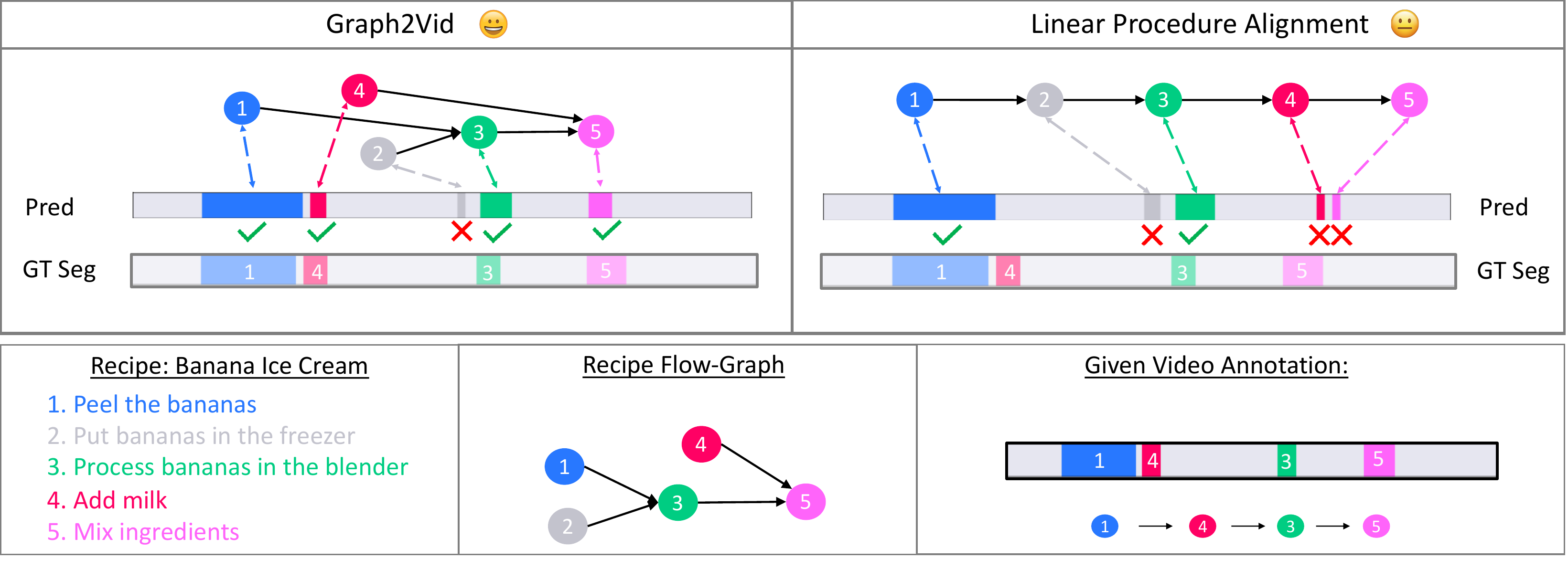}
	\caption{\textbf{Graph2Vid vs Linear Procedure Alignment for step localization on CrossTask.} Given a video of making Banana Ice Cream (bottom), we compare video segmentation (into steps) produced by both methods. Graph2Vid (top-left) localizes all the present recipe steps by grounding the flow-graph into the video, while aligning linear recipe sequence to the video (top-right) fails in 2 out of 4 steps. This is because the order of instruction in the video (bottom-right) is different from the linear recipe, however it conforms to the flow-graph and thus can be grounded by Graph2Vid. Both methods incorrectly predict the step ``2", that is actually not present in the video.
	}
	\label{fig:experiment}
	\vspace{-15pt}
\end{figure}

\subsection{Graph2Vid for step localization}\label{sec:results}


We begin by evaluating the proposed Graph2Vid technique as an inference time method for step localization in instructional videos. For this purpose, we follow previous work \cite{miech2020end,dropdtw} and rely on video and text features extracted from a model pre-trained on the HowTo100M dataset \cite{miech19howto100m}. First, the results in Tab~\ref{tab:inference} show better performance of the ``Linear Procedure" baseline compared to the ``Bag of Steps". This indicates that treating the recipe as a linear sequence of steps (\ie having some prior on the order of steps) allows for better step localization than treating the recipe as unordered set of steps. 
However, treating the recipe as a flow graph (automatically extracted using the learning-based parser of Sec.~\ref{sec:parser}) and grounding it in the video with Graph2Vid yields superior localization performance.
This confirms the advantage of using the more flexible graph structure of the recipe for step localization.
See Fig.~\ref{fig:experiment} for an illustration of how Graph2Vid takes advantage of the nonlinear flow-graph structure in an example from the dataset.
Unfortunately, there is still a large gap in performance between Graph2Vid and our upper bound, which relies on ground-truth ordered steps provided by human annotators. Closer examination of these results revealed that this gap is largely attributed to the fact that the GT of CrossTask does not conform to the assumed recipe flow graph structure, which we use in our Graph2Vid formulation. Specifically, while the flow graphs assume that each step happens once across the video, the GT of CrossTask allows for repeated steps (\eg \texttt{[cut tomato, cut cucumber, cut tomato, mix ingredients}]), where these repetitions are often a consequence of video post-production.

\begin{table}[t]
    \parbox[t][][t]{0.45\linewidth}{
	\caption{\textbf{Graph2Vid as an inference time procedure.}} 
	\resizebox{0.4\textwidth}{!}{
		\begin{tabular}{l| c c}
			\hline
			Inference Method & Acc.$\uparrow$ \ & IoU $\uparrow$ \\
			\hline
			Graph2Vid (ours) & \textbf{25.3}         & \textbf{17.1}      \\
			Linear Procedure~\cite{dropdtw}     & 22.3         & 15.1      \\
			Bag of Steps     & 20.5         & 13.7      \\
			\hline
			GT Step Sequence~\cite{dropdtw} & 32.4         & 21.2      \\
			
			\hline
			
		\end{tabular}\label{tab:inference}
	}
	}
	\hspace{0.02in}
    \parbox[t][][t]{0.35\linewidth}{
	\centering
	\caption{\textbf{Graph2Vid as a training loss.}}
		\begin{tabular}{l| c c}
			\hline
			Training Method & Acc. $\uparrow$ \ & IoU $\uparrow$ \\
			\hline
			Graph2Vid (ours)           & \textbf{26.3}           & \textbf{19.1}    \\
			Linear Procedure + Drop-DTW & 25.0           & 16.6    \\
			Bag of Steps + Soft Clustering  & 25.4           & 17.3    \\
			\hline
			Pre-trained Features~\cite{miech2020end}   & 25.3        & 17.1          \\
			\hline
			GT Step Seq. + Drop-DTW  & 35.7        & 25.3          \\
			\hline
			
		\end{tabular}\label{tab:training}
	}
\end{table}

\subsection{Graph2Vid for representation learning}

Here, we show the benefits of using the differentiable approximation of Graph2Vid for weakly-supervised representation learning. We start from the same video and text features used in Sec~\ref{sec:results} and train a two-layer multi-layer perceptron (\ie MLP) on top of the video representation.
During training, we assume no access to ground-truth ordered step sequences, but only to the task label (\eg making pizza) and its corresponding flow graph. The flow graph is obtained automatically from the procedure text description of the task using the learning-based graph parser descibed in Sec.~\ref{sec:parser}.
Here, we compare Graph2Vid for representation learning with two other methods: \textbf{(i)} procedure to video alignment with Drop-DTW~\cite{dropdtw} (\ie ``Linear Procedure + Drop-DTW"), and \textbf{(ii)} aligning video to a set of instruction steps using soft clustering (\ie ``Bag of Steps + Soft Clustering"). We elaborate on these baselines in Appendix.

Once again, the results in Tab.~\ref{tab:training} speak in favor of the proposed Graph2Vid approach, which better benefits from training compared to the considered baselines. Interestingly, training with ``Linear Procdure" does slightly worse than no training at all, indicating that aligning the video to a potentially out-of-order sequences, which is often the case for a fixed instruction list, results in a poor training signal.
In contrast, allowing Graph2Vid to infer the optimal topological sort of the flow-graph for aligning with the video results in a better training signal and improves video representations. Finally, using the GT ordered steps for training - a much richer source of supervision - yields best overall performance, however at the price of extra labeling effort.

\subsection{Ablation Study}\label{sec:ablation}

Here we evaluate the role of the flow graph parser and study the inference speed of our proposed Graph2Vid.\\
\textbf{Role of the flow graph parser.} To better understand the connection between the flow graph construction method and Graph2Vid localization performance, we compare the two different parsers described in Sec.~\ref{sec:parser} as well as graphs obtained from manual annotation.
As expected, the results summarized in Tab.~\ref{tab:parsers}, show that graphs from the rule-based parser yield performance slightly inferior to the manually generated graphs.
Surprisingly, the flow-graphs from the learning-based parser do better than the manually annotated flow-graphs on CrossTask.
After visually comparing the flow graphs produced by learning-based parser to the manual annotations, we realize that the former sometimes``misses" the edges present in the manual graph.
This essentially allows for more flexible step ordering when aligning to a video, which benefits step localization on the CrossTask dataset.

\begin{wraptable}{r}{0.4\textwidth} 
	\centering
	\caption{\textbf{Graph2Vid using different flow graph parsers.}
	}
	\resizebox{0.4\textwidth}{!}{
		\begin{tabular}{l| c c}
			\hline
			Method & Acc.$\uparrow$ \  & IoU $\uparrow$ \\
			\hline
			Manual Annotation          & 24.8         & 16.8      \\
			Rule-based Parser          & 24.3         & 16.3      \\
			Learning-based Parser      & \textbf{25.3}         & \textbf{17.1}      \\
			\hline
			
		\end{tabular}\label{tab:parsers}
	}
\end{wraptable}

This is because many CrossTask videos depict procedures that do not conform to the manually labeled flow graph, due to the post-editing of the videos. 
We provide more detailed analysis of this matter in Appendix.



\noindent\textbf{Evaluation of execution time.}
Here, we evaluate the speed of Graph2Vid for inference on CrossTask and compare it to running the brute force solution for flow graph grounding that considers all topological sorts explicitly (see Sec.~\ref{sec:naive}).
The flow graphs in CrossTask (obtained using the learning-based parser) have an average of $8.6$ nodes and $2.7$ separate threads. 
Such graphs, on average, produce $1,700$ topological sorts and generate tSort graphs, \S, with about $60$ nodes.
With such compact tSort graphs, Graph2Vid takes $\approx 57$ms to ground a flow graph to a video. In contrast, the brute-force procedure requires $\approx 3.2 * 10^4$ ms, which is almost $3$ order of magnitudes slower than Graph2Vid.
This result speaks decisively in favor of our Graph2Vid approach for flow-graph to video grounding and confirms our theoretical derivations in Sec~\ref{sec:complexity}.

\section{Conclusion}
\vspace{-5pt}
In summary, we introduced a new formulation for step localization in instructional videos using procedure flow graphs.
In particular, we proposed the novel task of flow graph to video grounding, which relies on task level procedure description to yield a step-wise segmentation of instructional videos. To this end, we rely on automatically generated \emph{task-level} procedure flow graphs for step localization instead of relying on manually annotated, \emph{per-video} step sequences. This effectively makes the proposed solution more scalable and practical.
To solve the task of flow graph to video alignment, we developed a new graph-alignment-based algorithm - Graph2Vid - that demonstrates superior localization performance and efficiency  
compared to baselines.
In addition, we could improve video representations by training with flow graphs as  supervisory signal and using Graph2Vid as a loss function.

We believe that flow graphs are a more natural and informative representation of procedural activities, compared to linear instructions. Moreover, procedure flow graphs are only needed at the task level, rather than on a per video bases. As such, we believe the proposed formulation to hold a great promise in minimizing labeling efforts and defines new avenues for further research.

\section*{Acknowledgement}
We thank Ran Zhang for the help with flow graph creation and processing.
\bibliographystyle{splncs04}
\bibliography{bibref}
\newpage
\appendix
\chapter*{Appendix}
\section{Summary}
Our supplemental material is organized as follows: Section~\ref{sec:tsort-graph} elaborates on our approach to construct tSort graphs from flow graphs and presents an efficient implementation to tackle this step of our Graph2Vid approach. We then present a detailed complexity analysis of Graph2Vid compared to the brute-force approach in Section~\ref{sec:complexity}.
In Section~\ref{sec:graph-parsers}, we present details on both the rule-based and learning-based parsers that we used to convert procedural text to flow graphs.
We elaborate on our experimental setup in Section~\ref{sec:exp-details}. 
Finally, in Section~\ref{sec:flowgraph}, we provide additional analysis of flow graphs and their influence on step localization performance.

\begin{algorithm}[t]
\small
	\caption{tSort-graph Construction\label{algo:tsort_sup}} 
	\begin{algorithmic}[1]
		\State \textbf{Inputs}: \G - flow graph, $s$ - sink node in \G
         \State $E_{tSort}$ = []
		 \Comment{\textit{init edge set of the tSort-graph}}
		 
		 \State $q$ = queue((s, set()))
		 \Comment{\textit{init BFS queue}}
		 
		 \While{q}
		 \State $v, F$ = $q$.pop() \Comment{\textit{active node v, set of visited front F}}
		 \State $P_v$ = get\_predecessors($v$, \G) \Comment{\textit{Get predecessors of v in G}}
		 
		 \For{$v_{new}$ in $P_v \cup F$}
		 \State $F_{new}$ = $P_v \cup F / \{v_{new}\}$
		 \\
		 \State is\_feasible = True
		 \For{$v_{F}$ in $F_{new}$}
		 \Comment{\textit{Checking for feasibility of this path}}
		 \If{lowest\_common\_ancestor(\G, $v_{new}$, $v_{F}$) = $v_{new}$}
		 \State is\_feasible = False
		 \EndIf \EndFor
		 \\
		 \If{is\_feasible = True} 
		 \State $q$.append(($v_{new}$, $F_{new}$)) \Comment{\textit{Add new node to the queue}}
		 \State $E_{tSort}$.add((($v, F$), ($v_{new}, F_{new}$))) \Comment{\textit{Add new edge to \S}} \label{alg:tsort_addedge}
		 
		 \EndIf \EndFor \EndWhile
        \State \S = build\_graph\_from\_edges($E_{tSort}$)
		 \Comment{\textit{build the tSort graph}}
        \State  \textbf{Output:} \S
	\end{algorithmic}
\end{algorithm}

\section{Efficient algorithm for tSort graph construction}\label{sec:tsort-graph}
In Section 3.3 of the main paper we presented a simple algorithm for tSort graph construction. Here, we further elaborate on tSort graph construction and present the more efficient procedure implementation (actually used in our work).
In Algorithm~\ref{algo:tsort_sup}, we present the algorithm used in our implementation to construct the tSort graphs, \S, given the flow graph, \G.
The algorithm uses Breadth First Search (BFS) traversals in \G, starting from the sink of the graph and following the edges of \G in the opposite direction (moving backwards to the root).
During the traversal, we build the tSort graph, \S, such that each node of \S is a tuple $(v, F)$, where $v \in V_{G}$ is a node in \G, and $F \subset V_{G}$ is the subset of nodes visited so far.
Specifically, during the BFS traversal, $v$ is the node that is currently being considered and it is referred to as the \textbf{``active node"}. On the other hand, $F$ is the set of nodes that have been visited by the BFS traversal on separate threads (\ie distinct from the thread containing the active node $v$) and they are collectively referred to as the \textbf{``front''}.

\begin{figure}[t]
\centering
\includegraphics[trim=0 220 20 0,clip,width=0.99\textwidth]{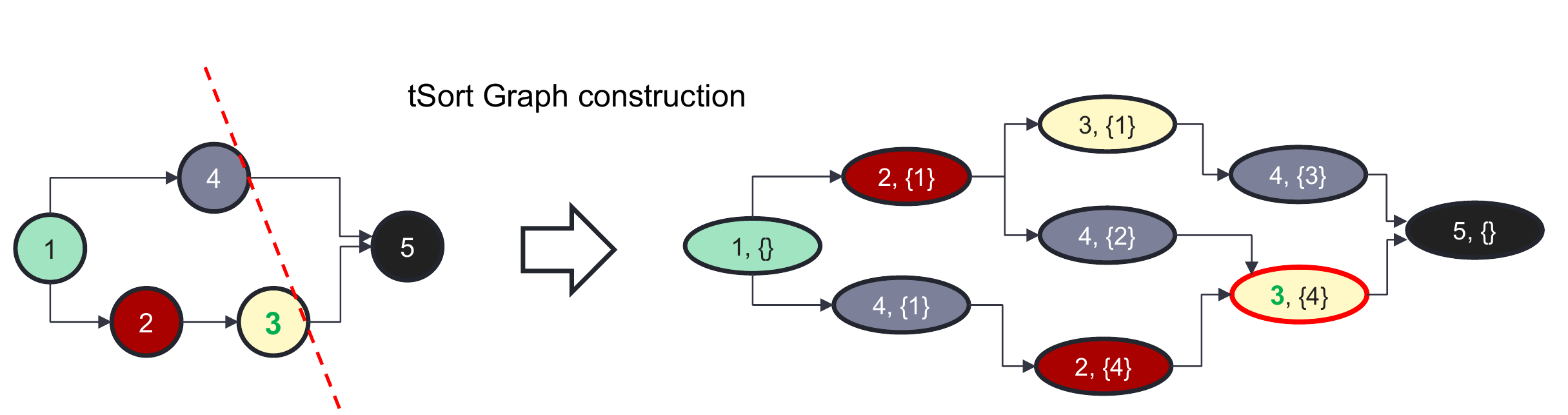}
\caption{\textbf{tSort graph construction.}
(left) Original flow-graph \G. The red dashed line illustrates the visited ``front" corresponding to the node of the tSort graph (3, \{4\}). (right) The obtained tSort graph \S. Every node in \S is a tuple, where the first element is the active node in \G, and the second is the ``front" of the traversal, i.e. the nodes last visited on all separate parallel threads. The node $(3, \{4\}) \in \mathcal{S}$ corresponds to the node $3 \in \mathcal{G}$ and the red dashed ``front" intersecting the parallel thread at node 4.
\label{fig:tsort_construct}
}
\vspace{-10pt}
\end{figure}

In the main paper, we explain that a tSort graph, \S, efficiently captures all the topological sorts of the graph, \G, while avoiding redundant paths. To achieve this property we check for path feasibility as described in Algorithm~\ref{algo:tsort_sup}. In particular, to make sure that during the BFS traversal we explore only the paths that conform to the original flow-graph, \G, (\ie valid topological sorts of \G), we check that the active node $v_{new}$ is \emph{not} an ancestor of any node in the front $F_{new}$ in the original graph, \G.
To understand why, remember that the set $F$ contains nodes already visited by the traversal, while $v_{new}$ is the node being visited currently. Therefore, if $v_{new}$ was an ancestor of one of the nodes $v_F \in F$ in \G, then $v_F$ must be visited \emph{after} $v_{new}$ in any valid topological sort of \G.
Thus, the transitions tuples $(v_{new}, F_{new})$ that do not satisfy the feasibility criteria above are not added to the traversal path (see Alg.~\ref{algo:tsort_sup}, lines 10-13).
In summary, using BFS for graph traversal (\ie an algorithm that is guaranteed to list all possible traversals of a graph) combined with the feasibility criteria described above, guarantees that \S contains \emph{all} the valid topological sorts of \G. Fig.~\ref{fig:tsort_construct} illustrates our tSort graph construction as described in Algorithm~\ref{algo:tsort_sup}.

\section{On the algorithm's complexity}\label{sec:complexity}

To develop some intuition on both the size of the generated tSort graphs, and on the speed-up over the naive approach described in Section 3.2 of the main paper, we consider simple model problems where the flow graph consists of $T$ separate, linearly-ordered threads, with $n_1, n_2, \ldots, n_T \geq 1$ nodes in each thread, for a total of $\sum_{t=1}^T n_t = n$ steps.  
For simplicity, we also add a unique root node, $s$, to \G, with edges to the beginning of each of the $T$ threads. We refer to such a flow graph as $\mbox{\G}(n_1, \ldots, n_T) = (V_G, E_G) $.  

For two threads ($T = 2$), ignoring the root node for a moment, the topological sorts of the flow graph are called riffle shuffle permutations, specifically $(n_1, n_2)$-shuffles~\cite{weibel1995introduction}.
These are analogous to the permutations that can be obtained from a sorted deck of $n$ cards by riffle shuffling a cut of the first $n_1$ with the remaining cards.
There are $N_{tSort}(n_1, n_2) = \binom{n}{n_1}$ such riffle shuffles. This analysis is easily extended to show that the number of topological sorts (tSorts) for our model problems are given by
\begin{equation}
    N_{tSort}(\mbox{\G}) = \frac{n!}{n_1!n_2! \ldots n_T!},~\mbox{where $n = \sum_{t=1}^T n_t$.}
    \label{eq:numTsort}
\end{equation}
Note that $N_{tSort}$ quickly becomes infeasibly large as $n$ and $T$ grow.

Next, for our model problems, we consider the number of nodes and edges in their tSort graph $\mbox{\S}(\mbox{\G}) = (V_S, E_S)$.  From Algorithm 1 we see that any node in $V_S$ is of the form $(v, F)$ where $v$ is the ``active'' node and $F$ is the set of all nodes last visited by the traversal on the separate parallel threads, referred to as the ``front''. For our model problems, $|F| \leq T-1$. 

This form $(v, F)$, clearly illustrates that, in this one node we are merging the prefix strings of all topological sorts that have arrived at node $v$ having processed nodes in $F$ {\tt in any other order}.  This merging of sequences to sets is the key to our efficiency gain.  

Moreover, for our model problems, we can use the form $(v, F)$ to
count the number of nodes, $|V_S|$, in \S, along with the maximum in-degree of edges in $E_S$.  Other than the root node, $(s, \emptyset)$, any node $(v, F)$ is defined by picking a thread, $t$, and an active node, $v$, from the $n_t$ nodes on that thread, and then forming $F$ to characterize the state of processing in the other $T-1$ threads.  Specifically, for these other threads, we might not have started on that thread (in which case we need $s \in F$), or we have already processed down to a specific node in that thread.  The total number of vertices is therefore seen to be
\begin{align}
    |V_S(\mbox{\G})| &= 1 + \sum_{t=1}^T \left[ n_t \prod_{j \neq t} (n_j + 1) \right],~\mbox{where $n = \sum_{t=1}^T n_t$,}
    \nonumber \\
    &= 1 +  \left[ \prod_{ j = 1}^T (n_j + 1) \right] \left[ \sum_{t=1}^T \frac{n_t}{n_t+1} \right]. \label{eq:numVerts} 
\end{align}
First, note that $|V_S|$ also rapidly grows with $n$ and $T$.  Therefore there will be practical limits to the size of flow graphs that are feasible to process in this manner, and in such cases we would need to resort to approximation approaches. However, in practice, we find that typical procedures result in tSort graphs of manageable sizes (see Section 4.5 of the main paper).  
Second, from Eq~\eqref{eq:numVerts}, we can note that the crude upper bound is $|V_S| = O(Tn^T)$. Hence, for a fixed number of threads $T$ in our model problems, the number of nodes in the tSort graph is polynomial in the number of nodes, $n$, in the original graph.  Finally, again for our model problems, the incoming edges at any node $(v, F)$ in the tSort graph must be due to a single step being performed in the flow graph, which must have occurred in one of the $T$ threads, either by advancing to the active state $v$ from the previous state, or by advancing to an element in the front $F$ in some other thread.  That is, there must be at most $T$ incoming (and, similarly, outgoing) edges to each node in \S.  

\begin{figure}[t]
\centering
\includegraphics[trim=0 0 0 0,clip,width=0.49\textwidth]{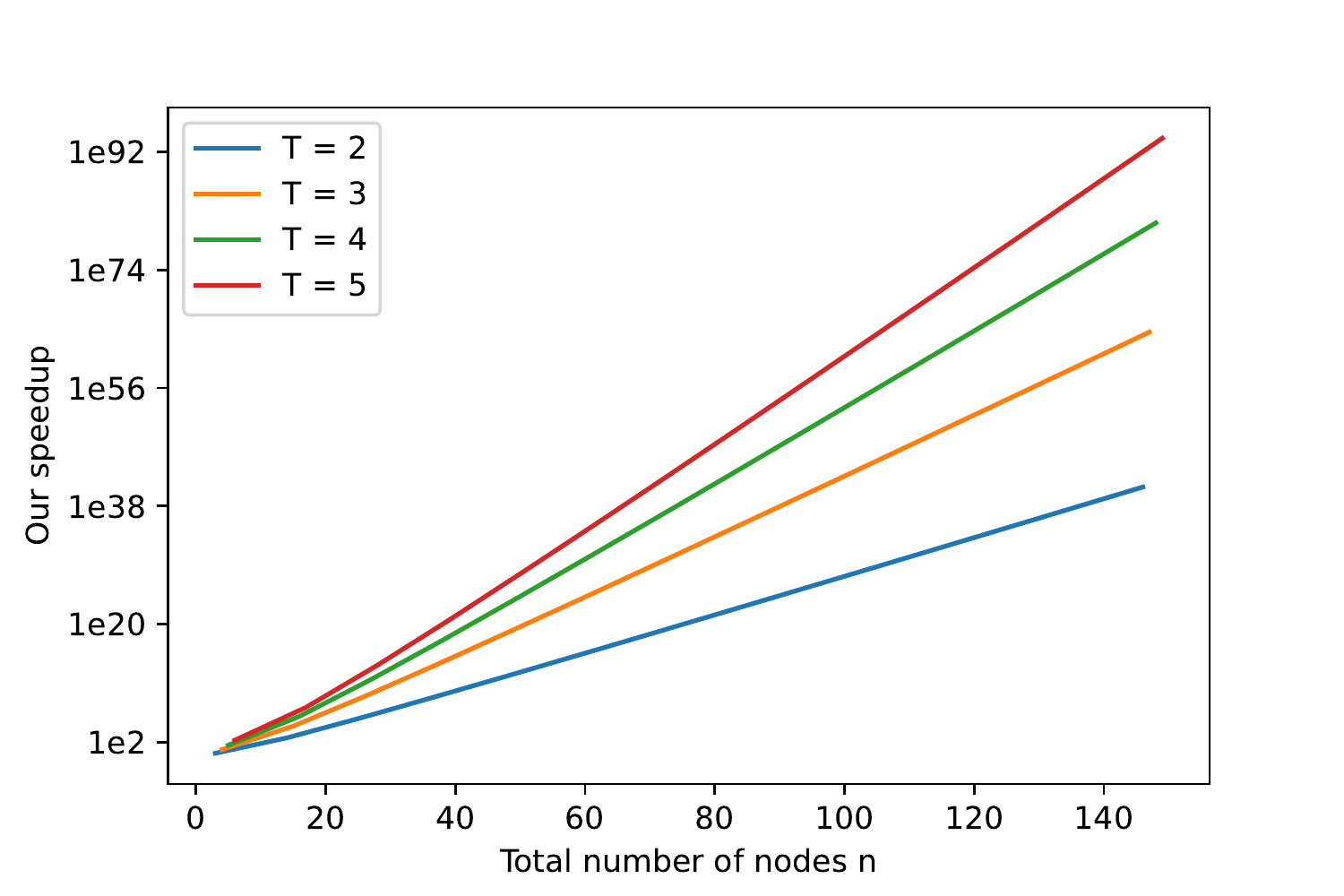}
\caption{\textbf{Complexity speedup with Graph2Vid over the brute-force approach.}
The plot shows gains in complexity (given by Eq.~\eqref{eq:complexity_ratio}) in log scale for the model example in Sec~\ref{sec:complexity} depending on the number of nodes $n$ in the flow graph, and for different number of threads $T$. 
\label{fig:complexity_speedup}
}
\end{figure}

The complexity of matching directly $N_{tSort}$ topological sorts of a flow graph, \G, with $|V_G|$ nodes to a video of $C$ clips, is $O(|V_G| N_{tSort}(\mbox{\G}) C)$, while the complexity of matching the associated tSort graph to $C$ clips is $O(T |V_S(\mbox{\G})| C)$.  The ratio of these leading order terms is therefore
\begin{equation}
\rho(\mbox{\G}) = \frac{ N_{tSort}(\mbox{\G}) |V_G| }{T |V_S(\mbox{\G})|}, \label{eq:complexity_ratio}
\end{equation}
and this ratio gives a rough idea of the speed-up. In Fig.~\ref{fig:complexity_speedup} we have plotted $log(\rho)$ as a function of $n$ for various numbers of threads, $T \geq 2$, for the above model problems, where the number of elements in each thread is $n_t \in \{\mbox{floor}(n/T), \mbox{ceil}(n/T) \}$. We can clearly observe massive speed-ups for our problem setup.

\section{Creating flow graph from procedural text}\label{sec:graph-parsers}
In this section we elaborate on the rule-based and learning-based graph parsers introduced in Section 3.6 of the main paper.
\subsection{Rule-based graph parsing}

\begin{figure}[t]
    \centering
    \includegraphics[width=\textwidth]{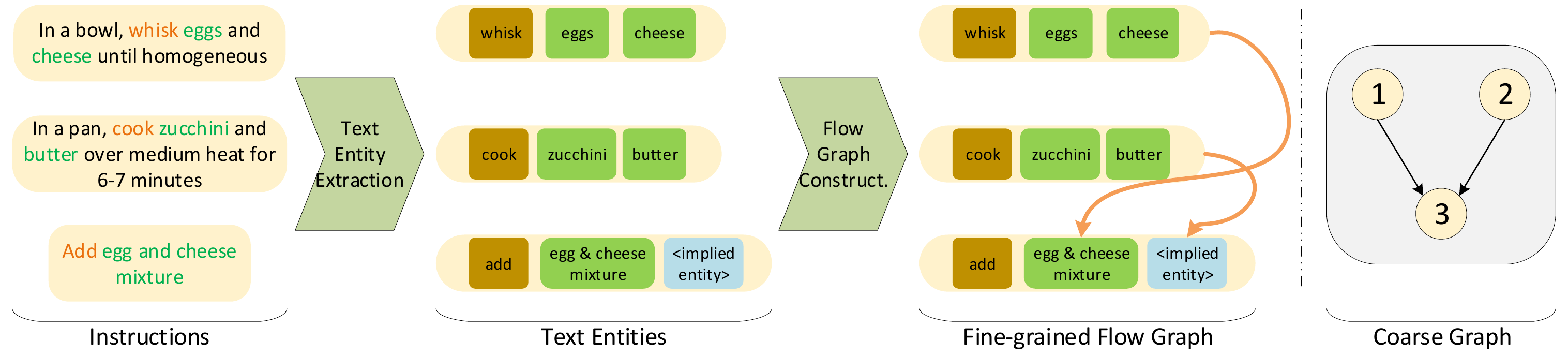}
    \caption{The text-to-flow graph generation pipeline. It consists of two major steps: extracting text entities from the original instructions, and constructing the fine-grained graph, in which objects and actions are properly connected to form a complete procedural flow graph. Finally, this fine-grained graph is collapsed into a coarse sentence-level graph used in our graph-to-sequence grounding algorithm.}
    \label{fig:graph_gen}
\end{figure}

Our flow graph construction pipeline is depicted in Figure~\ref{fig:graph_gen}. The pipeline consists of two main steps: text entity extraction and graph construction.

\noindent \textbf{Text entity extraction.} Starting from 
procedural text, we first extract relevant text entities from each individual sentence, including action verbs, direct and prepositional objects as suggested in~\cite{schumacher2012-pk}. Our entity extractor relies on an off-the-shelf dependency parser~\cite{spacy} in order to recover the 
verb and noun phrases from text. Furthermore, similar to previous work~\cite{Kiddon2015-ot}, we also take into account \textit{``implicit objects''}, which are only implied from the text. For example, in the third sentence in Figure~\ref{fig:graph_gen}, the \textit{``egg and cheese mixture''} is added to something that was omitted from the writing. One can deduce that this implicit object refers to the \textit{``cooked zucchini''}, product of the action described in the second sentence. We implemented a set of specific rules on top of dependency parsing to augment the extracted entities with implicit entities. For instance, in the above example, we use the rule \textit{``ADD [list of objects] TO [destination]''} to fill in the missing \textit{destination} with the \textit{``implicit object''}. These implied entities are also used in the graph construction. 

\begin{figure}[t]
    \centering
    \includegraphics[width=\textwidth]{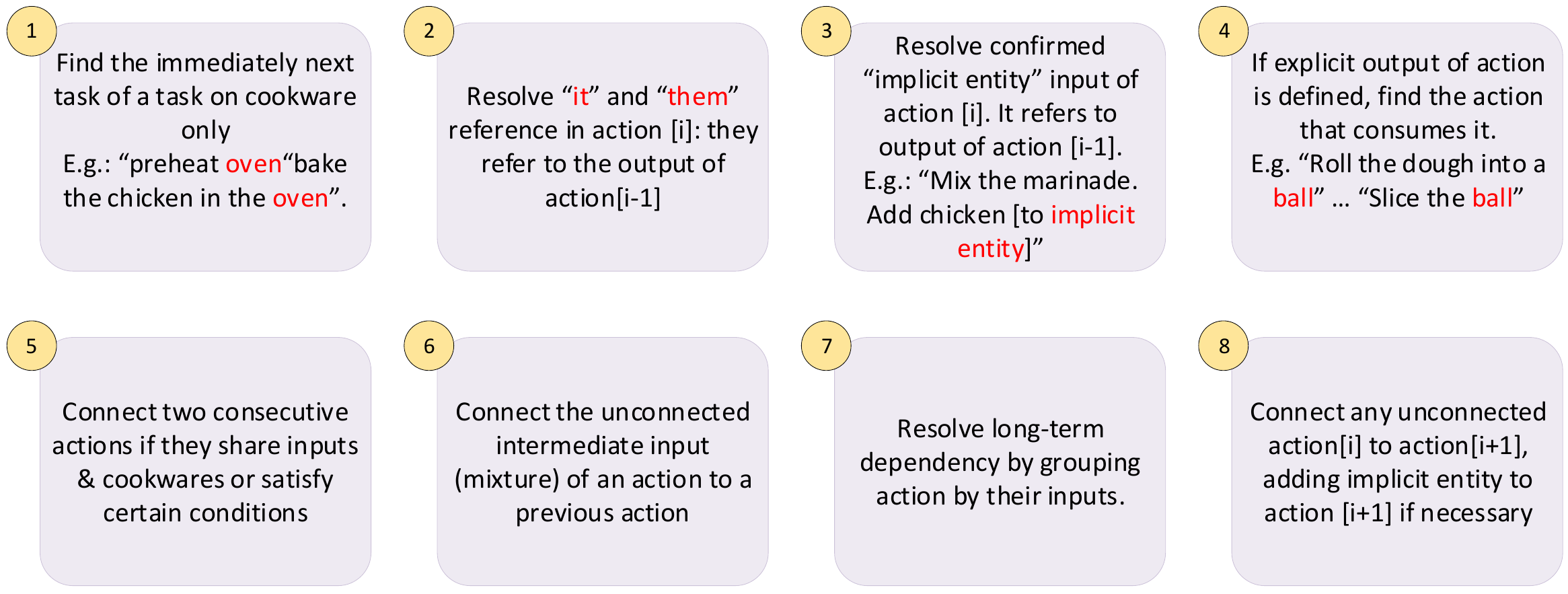}
    \caption{The rules used by our rule-based parser to convert procedure text into a flow-graph.}
    \label{fig:rules}
\end{figure}

\noindent \textbf{Flow graph construction.} Similar to previous work~\cite{Kiddon2015-ot,Yamakata2020-kk}, we assume that the output $p_i$ of an action $a_i$ in a graph is consumed by a subsequent action, $a_j$. In other words, one of the $K$ input objects $\{o_{jk}\}$ (including implicit entities) of $a_j$ is equivalent to $p_i$, $j$ indexes objects and $k$ indexes actions.  
To connect the various text entities, we defined a set of rich semantic rules for the graph constructor. The full set of rules with examples is given in Figure~\ref{fig:rules}.
Connecting the various entities with these rules results in a complete fine-grained flow graph as shown in Figure~\ref{fig:graph_gen}.
The fine-grained graph is then coarsened such that each node corresponds to a single instruction.
The resulting coarsened graph is what serves as input to Graph2Vid.

\subsection{Learning-based graph parsing}
The adopted learning-based flow graph construction follows the same two steps of the rule-based approach described above. However, in this case both entity extraction and graph construction yield from learning-based neural networks trained on the English recipe flow graph corpus~\cite{Yamakata2020-kk}.

\noindent \textbf{Text entity extraction.} Here, the text entity extraction is the output of the tagger model used in~\cite{donatelli2021}. In particular, this tagger is trained to recognize $10$ different named entities as done in previous work~\cite{Yamakata2020-kk}. For better accuracy, we re-trained the tagger on the Y-20 dataset~\cite{Yamakata2020-kk}. We used the Adam  optimizer with learning rate of $0.075$ and a batch size of $30$. We train the model for maximum of $100$ epochs with early stopping using accuracy measure on the validation set as a metric, and patience period of 10 epochs. We evaluate the quality of the tagger by computing precision, recall and F1 score. For all 10 tags, we get precision, recall and F1 of $0.87$, $0.88$ and $0.87$, respectively. 

\begin{figure}
    \centering
    \includegraphics[width=0.85\columnwidth]{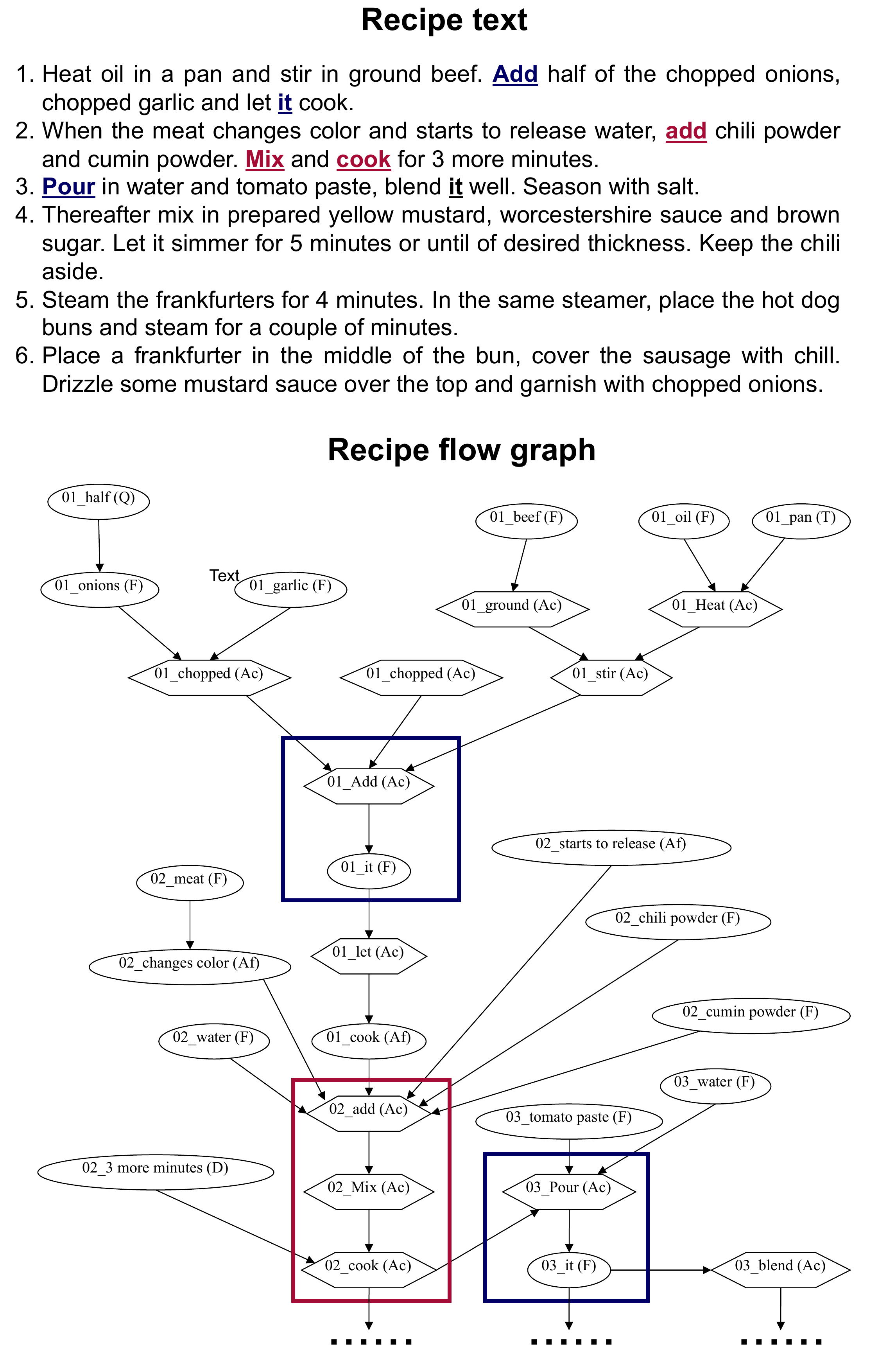}
    \caption{A recipe from our dataset, along with part of its learned flow graph corresponding to steps 1--3. Action nodes (e.g., add, mix) are shown as hexagons, while other entities (e.g., Food, Tool) are shown as ovals. Each node is labelled by an id (corresponding to instruction step), a token, and its entity class/tag (e.g., F for Food, Ac for Action). Colored rectangles identify instances of co-reference and ellipsis resolution in our parser, and will be further discussed in Section~\ref{sec:flowgraph}.}
    \label{fig:learned-fg}
\end{figure}

\noindent \textbf{Flow graph construction.} To construct the flow graph, we use the graph parser of~\cite{donatelli2021}, which takes as input the tagged entities from the previous step and converts them into a graph structure by predicting the presence of an edge between two entities, as well as a label indicating the semantic relation between them; see~\cite{Yamakata2020-kk} for details on entity and edge sets used by the parser.  Figure~\ref{fig:learned-fg} shows an example of a recipe and (part of) its corresponding flow graph learned by our parser. We evaluate the parser using standard measures, such as Unlabeled Attachment Score (UAS) and Labeled Attachment Score (LAS). UAS measures the number of nodes which are assigned correct parents, regardless of the edge label, while LAS takes into account correct edge label as well. Our parser yields UAS and LAS of $0.94$ and $0.91$, respectively. 

\vspace{10pt}
\subsection{Coarse flow graph generation}
In both the rule-based and learning-based approaches, the obtained fine-grained flow graphs are collapsed into coarse sentence-level graphs to be used in our experiments. To go from fine-grained (\ie entity-level) flow graphs to coarse (\ie sentence-level) flow graphs, we traverse the fine-grained graphs using Depth First Search (DFS) and merge all nodes with same sentence ID into a single node, while retaining original connections.

\begin{figure}
    \centering
    \includegraphics[width=\columnwidth]{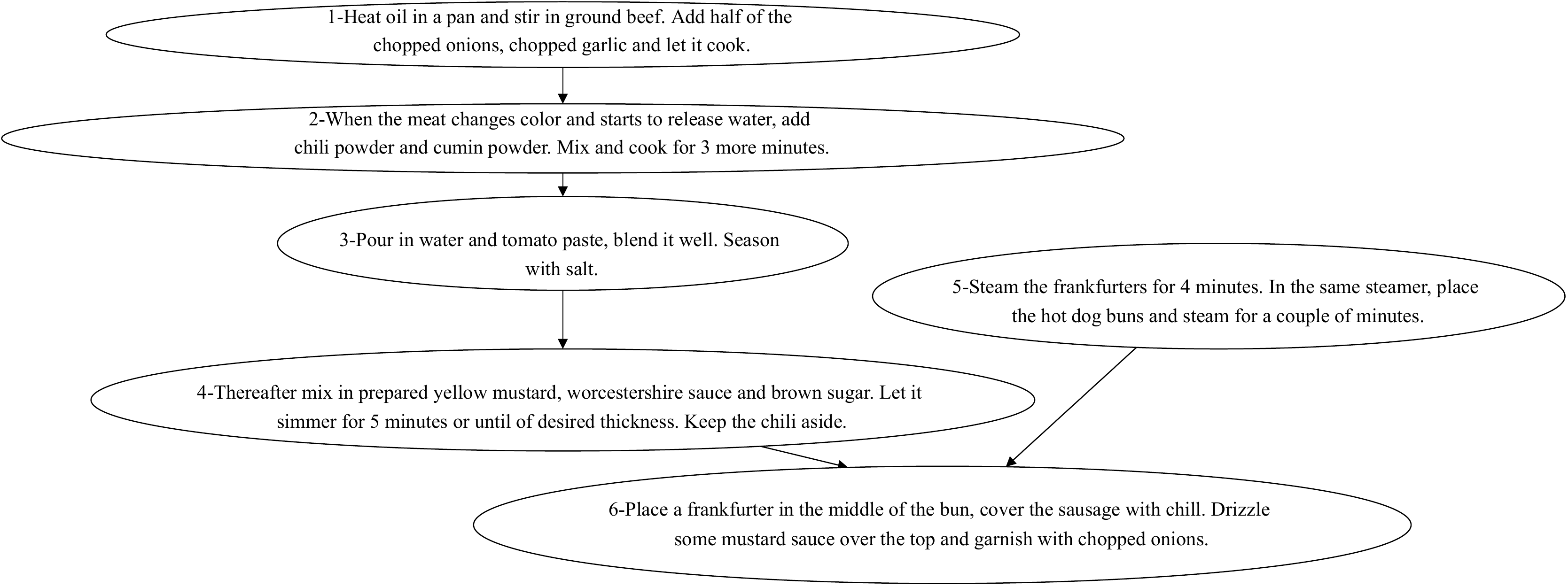}
    \caption{A coarse flow graph for recipe in Figure \ref{fig:learned-fg}. This is derived from a fine-grained recipe flow graph (part of it is shown in Figure \ref{fig:learned-fg}).  }
    \label{fig:coarse-fg}
\end{figure}





\section{Experimental}\label{sec:exp-details}
In this section, we detail the methods used for representation learning, presented in Table 2 of the main paper.

\textbf{Feature extraction.}
We start by elaborating on the feature extraction procedure.
Given a raw video $Y$ and a tSort graph $\mathcal{S}_{t} = (V_t, E)$ where every node $t_i \in V_t$ contains a step description in the form of text, we apply jointly pre-trained video ($f_v$) and text ($f_t$) feature extractors~\cite{miech2020end} (as mentioned in Sec.~4.3 of the main paper) to convert the video into a sequence of clip embeddings, $X = f_v(Y)$ and the text in each node into a sentence embedding $v_i = f_t(t_i)$.
All methods presented in Table 2 of the main paper expect such features as input.

\textbf{Pre-trained Features.} The baseline ``Pre-trained Features" in Table 2 of the main paper does not use any training at all.
Instead, we directly use the features from \cite{miech2020end} (as described above) and perform flow-grounding in videos using Graph2Vid.
In contrast to this method, all other approaches listed in Table 2 train a 2-layer Multi-Layer Perceptron (MLP) for the video features in order to improve step localization performance.

\textbf{Bag of steps + soft clustering.}
The following baseline assumes that the the recipe is an unordered set of steps (thus disregarding the connections in \G), and uses soft clustering loss to improve the pre-trained representations.
Given the sequence of video features, $X \in \mathbb{R}^{N \times d}$, and the list of step embeddings, $V \in \mathbb{R}^{K \times d}$, the soft clustering loss is defined as: %
\begin{align}
    \mathcal{L}_{\text{clust}} &= ||I - \hat{X}V^\top||_F, \label{eq:loss-clustering}
\end{align}
where $I \in \mathbb{R}^{K \times K}$ is the identity matrix and $\hat{X} = (\hat{x}_1, \dotsc, \hat{x}_K) \in \mathbb{R}^{K \times d}$. Each element $\hat{x}_i$ in $\hat{X}$ is defined according to
\begin{align}
    \hat{x}_i &= \sum_{j=1}^N x_j \cdot \textrm{softmax}(X v_i / \gamma). \label{eq:attn_pooling}
\end{align}
In other words, $\hat{x}_i$ in Eq.\ \eqref{eq:attn_pooling} defines attention-based pooling of sequence $x$, relative to an element $v_i$.
Minimizing $\mathcal{L}_{\text{clust}}$ pushes every element in $V$ to have a unique match in $X$, which encourages the clustering of the embeddings $x_i$ around the appropriate step embeddings $v_i$ and thus promotes relevant feature learning.
Note that this baseline does not use the knowledge of the flow graph structure or even the order of steps in $V$, since all the operations in soft clustering are permutation equivariant.

\textbf{Linear Procedure + Drop-DTW.}
In the following baseline, we treat instructions, as listed in the procedural text, as an ordered list of steps.
To train video features with step order supervision, we use Drop-DTW~\cite{dropdtw} and precisely follow their original implementation.
The only difference between our baseline and the original Drop-DTW~\cite{dropdtw} is the source of supervision. Specifically, the original work uses the provided ground-truth steps order, while we use the steps of the generic procedure (identical for all the videos of the same category) in the order in which they appear in the procedure description.
\\
To this end, given the video features $X \in \mathbb{R}^{N \times d}$, and the list of step embeddings, $V \in \mathbb{R}^{K \times d}$ we train the model with the following loss:
\begin{equation}
 \mathcal{L}_{train}(V, X) = \mathcal{L}_{DTW}(V, X) + \mathcal{L}_{clust}(V, X),
\end{equation}
Where $\mathcal{L}_{DTW}(X, V)$ is the cost of aligning the instruction list $V$ with the video $X$, and $\mathcal{L}_{clust}$ is the soft clustering loss (defined above) that~\cite{dropdtw} proposes to add to regularize the training.

\textbf{Graph2Vid.}
Finally, as described throughout our paper, Graph2Vid is a way to train video representations supervised by flow graphs, \G.
Please refer to Section 3.4 of the main paper for a detailed description of this proposed formulation.
Notably, to make the training with Graph2Vid more stable and avoid degenerate solutions where all the features map to a single graph node, we also adopt the regularization strategy from~\cite{dropdtw} and add the clustering loss. 
That is, our final training objective is
\begin{equation}
 \mathcal{L}_{train}(G, X) = \mathcal{L}_{G}(G, X) + \mathcal{L}_{clust}(V, X),
\end{equation}
where $\mathcal{L}_G(G, X)$ is the cost of aligning the flow graph, \G, with the video, $X$, and $\mathcal{L}_{clust}$ is defined above in Eq~\eqref{eq:loss-clustering}.

\textbf{Training details.}
In our experiment with Graph2Vid (as well as ``Linear Procedure + Drop-DTW" and ``Bag of steps + soft clustering" baselines) on CrossTask, we use a 2-layer MLP on top of the video features and train it with ADAM optimizer~\cite{kingma2014adam} with learning rate $10^{-4}$ and weight decay $10^{-4}$ for 10 epochs.

\begin{figure}[t]
\centering
\centering
	\includegraphics[trim=0 220 20 0,clip,width=\textwidth]{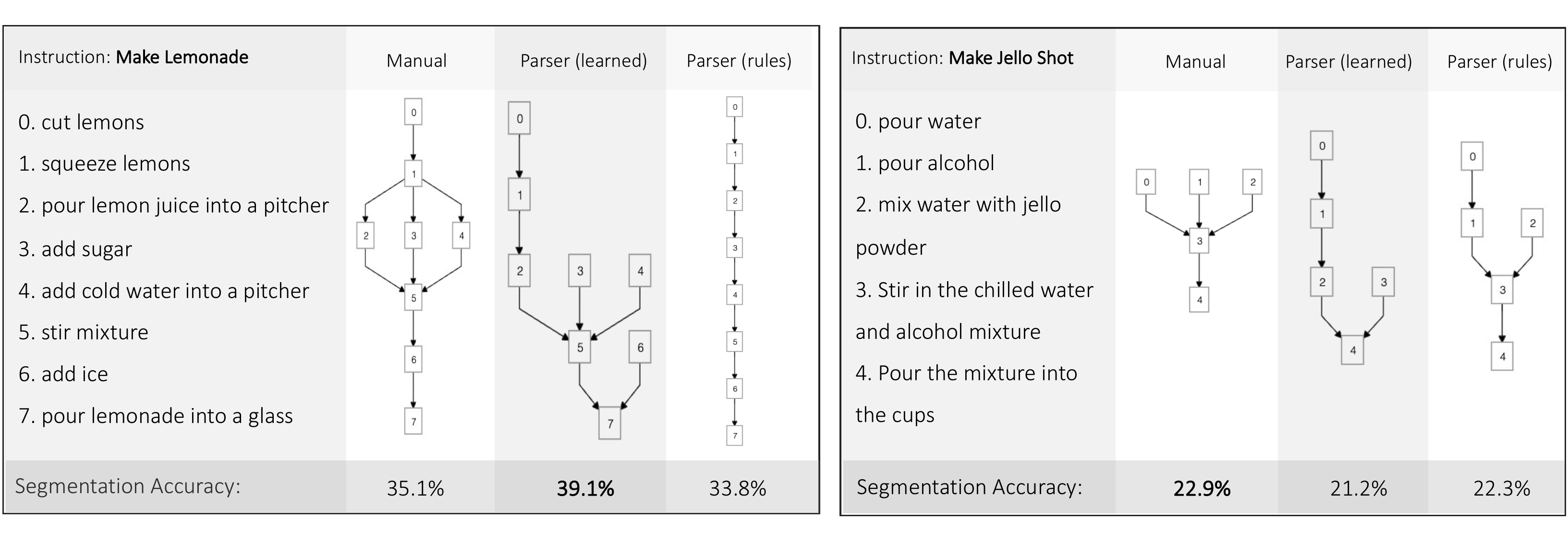}
	\caption{\textbf{Manual vs parser-generated flow graphs.} For two recipes, we provide the list of steps and derived from them flow graphs. The bottom row gives Graph2Vid segmentation accuracy on the videos of those procedures.}
	\label{fig:parsers}
\end{figure}

\section{Flow Graph Analysis}\label{sec:flowgraph}
In this section, we provide additional analysis of flow graphs and their influence on step localization performance.
\subsection{Comparison of different flow graphs for step localization}
To better understand the difference in step localization performance when using different flow graph, we compare manual and parser-generated flow graphs (see Fig.~\ref{fig:parsers}). On the “Make Lemonade” recipe (Fig.~\ref{fig:parsers}, left), the learning-based graph ignores some links present in the manual graph(e.g., 1 $\rightarrow$ 3, 1 $\rightarrow$ 4, and, 6 $\rightarrow$ 7) which leads to more parallel threads. On the other extreme, the rule-based graph detects a single linear order. This shows that the learning-based graph is more “flexible”, which we argue is key for better segmentation. Inspecting the CrossTask dataset videos revealed that the sequence of steps 5 $\rightarrow$ 6 $\rightarrow$ 7 (bottom of the manual graph) happens only 20\% of the time in the data, while the alternative $5 \rightarrow 7 \leftarrow 6$ (from the learned parser) happens 80\% of the time, which explains the higher learning-based segmentation accuracy (i.e., 39.1\% vs 35.1\%). Notably, sometimes both parsers produce more constrained flow graphs (e.g., Fig.~\ref{fig:parsers}, right), which leads to less accurate Graph2Vid segmentations.

\subsection{Co-reference and ellipsis resolution by the learned parser}
Here we provide an analysis of how our learned parser is capable of performing co-reference and ellipsis resolution. A quantitative analysis of the success rate of such resolutions requires a detailed manual annotation of the predicted flow graphs, and is beyond the scope of our study. 
But we observe many successful examples in our predicted flow graphs, which can be attributed to the way such resolutions are built into the annotation framework. Next, we present a few examples and elaborate on the reason behind the success of the parser in resolving them.
Recall that the flow graph annotations of~\cite{Yamakata2020-kk} draw on a set of pre-defined entity tags (including Action, Food, Tool, etc.), as well as a set of edge labels that identify relations among these entities (e.g., f-eq for food equivalency, t for target). These edge labels are used to connect nodes of different entity types. E.g., when f-eq connects two Food nodes, it signifies that the two Food items are equivalent (e.g., {\it potato} and {\it diced potato}). This same label can be used to connect an Action and a Food node, meaning that the result of the Action is the same as the Food node. 

The edge labels by design can help resolve the referent of pronouns that refer back to the result of a previous cooking action. We can see two such examples in the flow graph shown in Figure~\ref{fig:learned-fg} (inside the blue boxes), which result in correct resolution of references for the pronoun {\it it}. Similarly, the edge label t can connect an Action to its direct object (e.g., the connection from {\it heat} (Ac) to {\it oil} (F) in Figure~\ref{fig:learned-fg}), but also an Action to another Action whose result is the missing direct object of the first Action, and as such can help with ellipsis resolution.
Figure~\ref{fig:learned-fg} provides two
examples of such ellipsis resolution (inside the red box), where the result of {\it add} is identified as the implicit (missing) argument of {\it mix}, whose result is in turn linked to {\it cook} as its (missing) direct object.
All in all, because the training data contains many instances where co-reference and ellipsis cases are explicitly resolved via the use of proper edge connections and labels among entities, the parser is in principle capable of resolving them at inference time.
\end{document}